# A Model for General Intelligence

Paul Yaworsky
*Information Directorate*
*Air Force Research Laboratory*
*Rome, New York 13441*
November 14, 2018

## Abstract

The overarching problem in artificial intelligence (AI) is that we do not understand the intelligence process well enough to enable the development of adequate computational models. Much work has been done in AI over the years at lower levels, but a big part of what has been missing involves the high level, abstract, general nature of intelligence. We address this gap by developing a model for general intelligence. To accomplish this, we focus on three basic aspects of intelligence. First, we must realize the general order and nature of intelligence at a high level. Second, we must come to know what these realizations mean with respect to the overall intelligence process. Third, we must describe these realizations as clearly as possible. We propose a hierarchical model to help capture and exploit the order within intelligence. The underlying order involves patterns of signals that become organized, stored and activated in space and time. These patterns can be described using a simple, general hierarchy, with physical signals at the lowest level, information in the middle, and abstract signal representations at the top. This high level perspective provides a "big picture" that literally helps us "see" the intelligence process, thereby enabling fundamental realizations, a better understanding and clear descriptions of the intelligence process. The resulting model can be used to support all kinds of information processing across multiple levels of abstraction. As computer technology improves, and as cooperation increases between humans and computers, people will become more efficient and more productive in performing their information processing tasks.

## 1 Introduction

Intelligence is a complex process, but ultimately the process contains order. The order stems from the nature of the external world. The brain, or mind, contains its own kind of internal order. The basic premise here is that the internal model must contain useful representations of the external world. As part of the intelligence process, external order becomes internally realized and represented in the brain. But when you try to model intelligence in something other than the brain, say, in a computer, big problems arise. Ever since computers have existed, people have attempted to model intelligence using computers, with rather limited success.

A different perspective is needed to better realize fundamental aspects of intelligence. We propose a novel approach based on a simple, high level model. This perspective provides a general hierarchy that literally helps us "see" the intelligence process. Using this basic hierarchy as a guide, we gain a better understanding of intelligence, which in turn enables better descriptions of the process. Over time, this general model will enable specific implementations of intelligence using computer hardware and software.

One type of signal involved in intelligence is generically called information. Everyone is familiar with the term, yet no one seems to have a very clear understanding of how it plays into the intelligence process. Using this model, we will show how information and intelligence are interrelated. We will also discuss how information is simply one among many forms of signals, a means to an end, but not necessarily the most significant or most useful form of signal.

## 2  Background

Efforts to understand the workings of the mind go back thousands of years. But it was not until the advent of the electronic computer, however, that lots of people got really interested in the notion of modeling intelligence using something other than the human brain (mind). As the computer industry began to take off, very many disciplines and sub-disciplines of Artificial Intelligence (AI) sprang up, each aimed at modeling a portion of intelligence. Some generic terms used today which fall under the umbrella of AI include neural networks, cognitive computing, machine intelligence, machine learning and deep learning.

Over the past sixty years, countless efforts and amounts of money have been spent trying to solve problems associated with modeling intelligence using computers. In 1956, the Dartmouth Summer Research Project on Artificial Intelligence kicked off the field of AI. In their proposal [8] for the famous Dartmouth conference, the founding fathers of AI wanted to "make machines use language, form abstractions and concepts." These concepts are fundamental to AI and vital to the model described here.

Attempts to engineer the intelligence process depend on making a set of basic realizations. However, these realizations have proven to be elusive and counter-intuitive, since they involve externalizing that which is internal and formalizing that which is abstract. In his research on vision, perception and pattern recognition, David Marr described "Three Levels" of analysis and understanding [12]. His philosophy and approach encompassed much more than the problem of vision though. In general, Marr's "Three Levels" are: 1) What problem are you trying to solve; 2) How will you solve the problem; and 3) Solve the problem using a physical implementation. Marr pointed out that without properly addressing issues at Level 1, work done at Levels 2 and 3 may not be well-founded. This challenge, as posed at Marr's Level 1, has been at the root of countless problems for AI researchers for more than sixty years.

More recently, there has been a surge of activity and progress in AI. One example is IBM Watson [9]. According to IBM, their Watson system is a cognitive computer, and they describe cognitive as being able to think, learn, understand, reason and use natural language. The 2011 *Jeopardy!* Watson Challenge was a significant achievement in its own right. But that was a very structured event operating within tight constraints, even though one of the difficult constraints was human language (text) processing. Since then, IBM has further developed and enhanced their system, but we must realize that even today, Watson's performance is nowhere near that of human level with respect to general intelligence.

Another approach to AI can be found in the field of machine intelligence. For example, Jeff Hawkins [5] and the folks at Numenta aim to discover the operating principles of the human neocortex and to develop machine intelligence technology based on those principles. Hawkins has described intelligence in terms of patterns that become organized into spatio-temporal hierarchies in the brain. He has also pointed out something that Francis Crick mentioned earlier, that a "broad framework of ideas" is needed to help describe the intelligence process. The work described in this paper aims to provide such a framework as we attempt to capture the essence of the intelligence process in terms of a general, high level model.

Google is also heavily invested in AI. Their mission [7], to "organize the world's information and make it universally accessible and useful," is basically an AI mission, since it combines what intelligence can do with what vast amounts of interconnected computers can do. Google DeepMind has a mission even more focused on intelligence [6], and that is to "solve intelligence and use it to make the world a better place." Their approach incorporates deep neural networks, which offer the benefits of learning, layers, levels, abstractions, hierarchies, patterns, memory and representation, among all that is needed to solve the intelligence problem.

Other large US companies such as Amazon, Facebook, Apple, Intel, NVidia and Microsoft are interested in AI as well. Of course, countless organizations large and small, all over the world, have also become focused on AI. While much progress has been made in decades past and more recently, it seems like more questions may have been raised than answered [11]. Perhaps the right questions have not yet been asked. An underlying problem is that the intelligence process is not understood well enough to enable sufficient hardware or software models, to say the least.

We take a fundamental look at the nature of the intelligence process. At a high level, this kind of work is abstract and general by its very nature. But we must realize that two of the most fundamental functions a neuron can perform are to abstract and to generalize. Neurons do this continuously as they process many inputs and produce a single output (that is, neurons make decisions for a living). And when you scale up this functionality to higher and higher levels of intelligence, you get more significant abstractions and generalizations. You also get hierarchies. And along the way, 'information' just has to play an important role in the process. These fundamental concepts are essential to our approach in the development of a general model for AI.

## 3  A Simple, Hierarchical Model

The overarching problem in AI is this: what is the nature of the intelligence process such that we can model it using computers? To solve this problem, we focus on three general aspects of intelligence. First, we must realize (to the extent possible) the general nature of intelligence. Second, we must know what these realizations mean or represent with respect to the overall intelligence process. Third, we must be able to describe these realizations as clearly as possible. Note that initially these descriptions can come in any form, using any language (for example, English). Inherently, any description will necessarily involve the concept of information. All this will result in a basic understanding of intelligence. We must then describe this understanding clearly enough to enable subsequent implementations in computer hardware and software. The resulting model may serve as a foundation to support future work incorporating the functionality of intelligence combined with the benefits of computer technology.

In this paper we portray intelligence as an orderly, organized process using a simple hierarchy (Figure 1). With three basic levels of realization and with information at the core, this hierarchy offers a unique perspective of the intelligence process. The following description provides a brief overview of this general model.

The real world naturally contains order and organization. An important job of the human brain is to try and exploit that order and organization to some extent, thereby forming an internal model of the external world. This model enables learning and understanding, which further enable internal processes such as memory and decisions, as well as external activity such as movement and communication. The overall flow of input signal activity and learning within the model is bottom-up, with a many-to-one kind of reduction process occurring. The overall flow of output activity, or recall of learned signal representations, tends to be top-down. Understanding is a top-down/bottom-up harmony or resonance of signal activity. Intelligence is the ability to learn and recall effectively in a changing environment.

Physical activity occurring in the external world may be perceived by the senses. Resulting signals are transmitted to the brain. Over time, with experience and through learning, these signals become transitioned, transformed and stored as rich sets of signal patterns. These patterns are dynamic and considered very complex. However, fundamentally they must be the combination of many simple components. One useful feature of these patterns is that their components contain order, and that order can be represented using a general intelligence hierarchy.

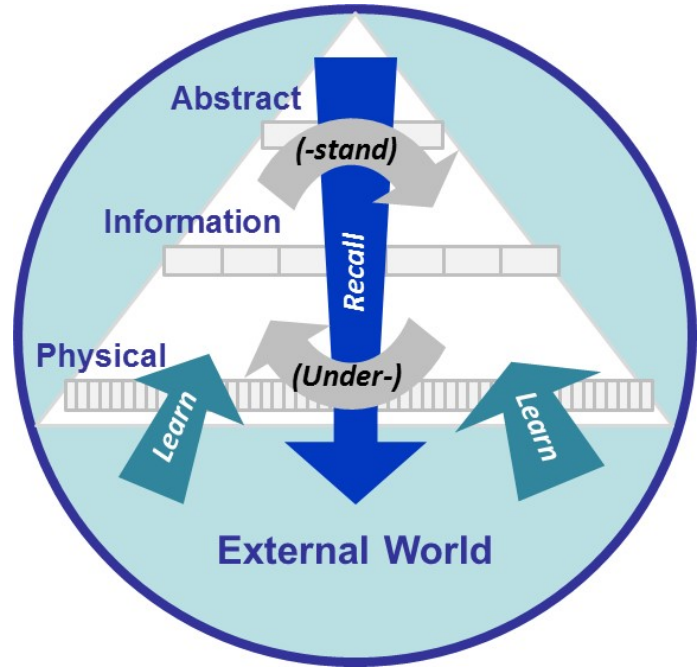

Figure 1. General intelligence hierarchy.

Nature's external order and organization can be modeled internally in the brain (mind) in the form of very many hierarchies. The brain uses neurons and connections as basic building blocks to form these hierarchies. At a small scale, each neuron takes many inputs and generates a single output. Out of all the activity, through experience and learning, connections among neurons are formed. Over time, very many neurons and connections combine and scale upward to form complex hierarchies across multiple layers of processing. These hierarchies capture and represent the order present in signals coming from the external world. This externally driven activity combines with internally existing patterns, resulting in a general model (i.e., mind) that represents all the signal activity occurring within the brain. Along with hierarchies comes rank ordering, with the most significant things at the top, relatively speaking. Next we discuss hierarchies in a temporal as well as spatial sense. Then we combine these two kinds of hierarchies to form a simple, single model for general intelligence.

## *3.1 Temporal Hierarchy*

In a temporal sense, external inputs to the brain contain signals that are dynamic; that is, the signals have temporal components that change over time. Our senses perceive signals as having timely or time-sensitive characteristics. Such characteristics include signals with short intervals or short time scales associated with them (relatively speaking). These signals are real-world, dynamic and continuously changing. However, beginning with all this so-called low level activity, signals travel up the hierarchies and transition through multiple levels. Along the way, *time* gets abstracted out of the signals. Ultimately, all that remains is their essence. What this means is that high level signal entities (representations) tend toward being timeless; they have characteristics which are slow-changing, have a long time scale associated with them, or they are non-changing. These include signal representations such as general knowledge, basic principles and fundamental theories. Granted, the temporal characteristics associated with these kinds of signals are relative, but nonetheless timing is one of their fundamental properties.

The temporal hierarchy can also be described using the so-called *data to information to knowledge* progression. In a temporal context, lots of low level data enters the brain via the senses. Input data tends to be quickly changing, time-sensitive and the least organized, relatively speaking. As signals travel up the hierarchy, they transition into information, an intermediate, more organized (albeit ambiguous) state. Then, at the highest level of the hierarchy, even fewer entities remain. These signals are the most significant with respect to time, a key example being knowledge. Relatively speaking, knowledge is time-insensitive, slow to change (if it changes at all) and the most organized of the three types of signal-states mentioned. As such, knowledge (cognition) is a significant part of the overall intelligence process.

## *3.2 Spatial Hierarchy*

Similarly, in a spatial sense, signals undergo transformations or changes with respect to form. These signals form hierarchical patterns that are orderly and organized spatially. Stemming from the external world, signals which have physical or formal characteristics may be perceived by our senses. This allows these formal, physical signals (e.g., images, symbols) to become represented in our brain. However, upon entering the brain, the signals go through various kinds of changes (transformations) in a spatial sense. As signals travel up the hierarchies, *form* gets abstracted out. Eventually, all that remains is their essence. What this means is that the resulting high level signal entities have *no form*. This includes things such as ideas, concepts and beliefs. And since signals naturally progress inward and upward, toward the top of hierarchies, this implies that higher level signals are more significant than the lower level signals used to form them. Furthermore, the spatial abstraction hierarchy reveals the importance of realization in producing these higher level entities. That is, lower level signals may originate in the real world in a physical, formal or concrete sense. But as they travel up the hierarchy, these signals become realized internally in different forms (mental representations). As such, the resulting representations become more important, and their realizations (abstractions) become essential components of high level intelligence.

## *3.3 General Hierarchy*

In a modeling sense, the temporal and spatial abstraction hierarchies may be combined to form a single, general hierarchy (Figure 1). This provides a big picture of the overall intelligence process. From a temporal point of view, the *data to information to knowledge* progression provides useful ways to characterize fundamental features of signal components with respect to time. From a spatial point of view, specific physical signals occurring at lower levels give way to more general, abstract entities at higher levels. In general, the overall flow of input signals is many-to-one, as signals travel upward within spatio-temporal hierarchies. This signal reduction process results in fewer and fewer entities near the top of hierarchies. Along the way, inherent signal characteristics become represented in fewer, more abstract and general ways. The higher level signal components are also more significant with respect to intelligence. Taken together, this orderly activity results in the realization of a model for general intelligence which captures hierarchical features of signals occurring and changing in both space and in time.

## 4 Impact, Levels, Concerns

What is the significance and impact of developing a model for general intelligence? More generically, so what if 'artificial intelligence' (we use the term loosely here) succeeds? What difference will it make? The short answer is that everyone will benefit. In the context of human-kind, an initial payoff is a raised level of understanding about intelligence; we can all learn more about how our mind/brain works in simple, general terms. In the context of computer technology, computers are becoming more and more capable. These days, we hear terms such as "smart" or "cognitive" or "intelligent" often used to describe the latest capabilities of digital devices (i.e., computer technology). The technology payoff is this: as computers continue to improve, and as cooperation increases between humans and computers, *people will become more efficient and more productive in their information processing activities*. We must remember that computers work for us! The field of artificial intelligence has been around for over sixty years, and much progress has been accomplished. However, it is clear that AI needs a better foundation and we have a long way to go. The model described here can be used to help support future generations of artificial intelligence.

Since natural intelligence involves information processing and signal activity at multiple levels of abstraction, representation and realization, it stands to reason that artificial intelligence might also incorporate similar levels of abstraction, representation and realization. With respect to the intelligence process, these levels can generally be described as follows (after Marr's Levels):

> Level 1) Understanding (abstract concepts, general knowledge, theory)
> Level 2) Description (information, algorithms, software, language)
> Level 3) Implementation (physical system, hardware, raw data)

However, the vast majority of work in AI has occurred at Levels 2 and 3. While the importance of hardware, software, algorithms, big data, etc. is obvious, we must also acknowledge that all levels of processing are needed to achieve AI. The approach described in this paper focuses mainly on Levels 1 and 2, with the implication that resulting insight, understanding and descriptions may be used to support all levels of intelligent information processing. Having a big picture is useful. Furthermore, it is very powerful when top-down and bottom-up approaches can effectively be combined.

What about the concerns of AI running amok and taking over human-kind? It is believed that AI will someday become a very powerful technology. But as with any new technology or capability, problems tend to crop up. Especially with respect to general AI, or artificial general intelligence (AGI), there is tremendous potential, for both good and bad. We will not get into all the hype and speculation here, but suffice it to say that many of the problems we hear about today concerning AI are due to sketchy predictions involving intelligence. Not only is it difficult to make good scientific predictions in general, but when the science in question involves intelligence itself, as it does with AI, then the predictions are almost impossible to make correctly. Again, the main reason is because we do not understand intelligence well enough to enable accurate predictions. In any event, what we must do with AI is proceed with caution.

## 5 Conclusion

Intelligence involves information processing at multiple levels of abstraction. Human intelligence no doubt involves high levels of processing (i.e., abstract, general), yet the majority of work in AI is done at more specific, lower levels of abstraction. Something (actually, a lot) is missing in AI. We address this void by developing a high level model for general intelligence. This model is hierarchical in nature, exploiting order present within the intelligence process. The order can be described using a simple, general hierarchy, with physical signals at the lowest level, information in the middle, and abstract signal representations at the top. This high level view provides a "big picture" of intelligence, thereby enabling fundamental realizations, better understanding and more clear descriptions of the process. Along with improvements in computer technology, including enhanced hardware, the incorporation of learning and better methods of cooperation, humans will come to use their computational counterparts in countless ways as we face the future together.

# References

1. Arbib, M., Editor, The Handbook of Brain Theory and Neural Networks, MIT Press, Cambridge, MA, 1995.
2. Churchland, P. and Sejnowski, T., The Computational Brain, MIT Press, Cambridge, MA, 1993.
3. Crevier, D., AI: The Tumultuous History of the Search for Artificial Intelligence, Basic Books, NY, 1993.
4. Fogel, D., Evolutionary Computation: Toward a New Philosophy of Machine Intelligence, IEEE Press, Piscataway, NJ, 1995.
5. Hawkins, J. and Blakeslee, S., On Intelligence, Owl Books, New York, NY, 2004.
6. https://deepmind.com/about/, March 2018.
7. https://google.com/about/our-company/, March 2018.
8. https://rockfound.rockarch.org/documents/20181/35639/AI.pdf/a6db3ab9-0f2a-4ba0-8c28-beab66b2c062, Proposal for the 1956 Dartmouth AI Conference, March 2018.
9. https://www.ibm.com/watson/about/, March 2018.
10. Kurzweil, R., The Age of Intelligent Machines, MIT Press, Cambridge, MA, 1990.
11. Marcus, G., "Deep Learning: A Critical Appraisal," arXiv, cs.AI, 2018.
12. Marr, D., Vision: A Computational Investigation into the Human Representation and Processing of Visual Information, W. H. Freeman and Company, San Francisco, CA, 1982.
13. Minsky, M., The Society of Mind, Simon and Schuster, NY, 1986.
14. Newell, A., Unified Theories of Cognition, Harvard University Press, Cambridge, MA, 1990.
15. Nilsson, N., Artificial Intelligence: A New Synthesis, Morgan Kaufman Publishers, Inc., San Francisco, CA, 1998.
16. Penrose, R., The Emperor's New Mind: Concerning Computers, Minds and the Laws of Physics, Oxford University Press, NY, 1989.
17. Pfeifer, R. and Scheier, C., Understanding Intelligence, MIT Press, Cambridge, MA, 1999.
18. Russell, S. and Norvig, P., Artificial Intelligence: A Modern Approach, Prentice Hall, Englewood Cliffs, NJ, 1995.
19. Simon, H., The Sciences of the Artificial, 3rd Edition, MIT Press, Cambridge, MA, 1996.
20. Sun, R., Coward, L. A. and Zenzen, M. J., "On Levels of Cognitive Modeling," Philosophical Psychology, Vol. 18, No. 5, pp. 613-637, 2005.
21. Yaworsky, P., "Realizing Intelligence," arXiv, cs.AI, 2018.
22. Yaworsky, P., "Cognitive Information Processing in a Sensor Network," Final Report to AFOSR, 2009.
23. Yaworsky, P., Holzhauer, D., Fleishauer, R. and Koziarz, W., "Cognitive Network for Atmospheric Sensing," Proceedings of the 2007 International Conference on Artificial Intelligence, pp. 427-432, CSREA Press, Athens, GA, 2007.
24. Yaworsky, P., "Framework for Modeling the Cognitive Process," Proceedings of the 10[th] ICCRTS, Washington, DC, 2005.
25. Yaworsky, P., "Toward Automating Intelligent Information Processing," Air Force Rome Laboratory Report, RL-TR-97-46, 1997.
26. Yaworsky, P. and Vaccaro, J., "Neural Networks, Reliability and Data Analysis," Air Force Rome Laboratory Report, RL-TR-93-5, 1993.